\documentclass[10pt,twocolumn,letterpaper]{article}

\usepackage{cvpr}              % To produce the CAMERA-READY version
\usepackage[utf8]{inputenc} % allow utf-8 input
\usepackage[T1]{fontenc}    % use 8-bit T1 fonts
\usepackage{url}            % simple URL typesetting
\usepackage{nicefrac}       % compact symbols for 1/2, etc.
\usepackage{microtype}      % microtypography
\usepackage[usenames, dvipsnames]{xcolor}         % colors
\definecolor{lightgray}{rgb}{0.9, 0.9, 0.9}
\usepackage{amsmath,amsfonts,bm}

\usepackage{booktabs}   % professional tables
\usepackage{tabularx}   % adjustable-width columns
\usepackage{colortbl}   % for coloring tables
\usepackage{multirow}
\usepackage{makecell}

\usepackage{mwe} % for placeholder images

\usepackage[font=normal,labelfont=bf]{caption}
\usepackage[font=normal]{subcaption}

\usepackage[normalem]{ulem} % underlining
\usepackage{array}

\usepackage{xparse}
\usepackage{pifont}

\usepackage{ifthen}

\definecolor{cvprblue}{rgb}{0.21,0.49,0.74}
\usepackage{algpseudocode}
\usepackage[linesnumbered,ruled,vlined]{algorithm2e}

\SetCommentSty{mycommfont}
\SetKwInput{KwInput}{Input}

\usepackage{enumitem}

\usepackage{xfp}

\PassOptionsToPackage{square,sort,comma,numbers}{natbib}
\definecolor{Gray}{gray}{0.93}
\newcolumntype{a}{>{\columncolor{Gray}}c}

\newcommand{\colorrow}{\rowcolor{blue!5}}

\newcommand{\inc}[1]{\textsuperscript{\tiny\textcolor{green!60!black}{$\uparrow$#1}}}
\newcommand{\incr}[1]{\textsuperscript{\tiny\textcolor{red}{$\uparrow$#1}}}
\newcommand{\dec}[1]{\textsuperscript{\tiny\textcolor{red}{$\downarrow$#1}}}
\newcommand{\decg}[1]{\textsuperscript{\tiny\textcolor{green!60!black}{$\downarrow$#1}}}

\usepackage{xspace} % for abbreviation macros
\makeatletter
\DeclareRobustCommand\onedot{\futurelet\@let@token\@onedot}
\def\@onedot{\ifx\@let@token.\else.\null\fi\xspace}

\makeatother

\definecolor{cvprblue}{rgb}{0.21,0.49,0.74}
\usepackage[pagebackref,breaklinks,colorlinks,allcolors=cvprblue]{hyperref}

\def\paperID{*****} % *** Enter the Paper ID here
\def\confName{CVPR}
\def\confYear{2026}

\title{Data Warmup: Complexity-Aware Curricula for Efficient Diffusion Training}

\author{
Jinhong Lin$^{1}$\thanks{Equal contribution.}\;\thanks{Corresponding author.} \quad 
Pan Wang$^{2}$\footnotemark[1] \quad 
Zitong Zhan$^{3}$ \quad
Lin Zhang$^{1}$ \quad
Pedro Morgado$^{1}$\\
\\
$^{1}$University of Wisconsin–Madison \quad
$^{2}$University of Pittsburgh \quad
$^{3}$University at Buffalo, SUNY\\
{\tt\small\{jlin398,lzhang756\}@wisc.edu, paw101@pitt.edu, zitongz@sairlab.org, p.morgado.89@gmail.com}
}

\begin{document}
\maketitle

\newcommand{\MainExp}{
% \begin{table}
\begin{tabular}{llllll}
    \toprule
    \textbf{} & \bf IS ↑               & \bf FID ↓               & \bf sFID ↓             & \bf Pre. ↑            & \bf Rec.  ↑           \\
    \midrule
    SiT-B/2~\citep{ma2024sit}  & 41.40 & 36.16 & 6.80 & 0.52 & 0.63 \\
    \colorrow{}%
    + Data Warmup 
    & 45.70\inc{4.30} & 32.75\decg{3.41} & 6.56\decg{0.24} & 0.54\inc{0.02} & 0.63 \\
    \hline
    \textcolor{gray}{+ Inverse Data Warmup}
    & \textcolor{gray}{36.60}\dec{4.80} & \textcolor{gray}{41.05}\incr{4.89} & \textcolor{gray}{7.19}\incr{0.39} & \textcolor{gray}{0.49}\dec{0.03} & \textcolor{gray}{0.62}\dec{0.01} \\
    \bottomrule
\end{tabular}
% \end{table}
}

\newcommand{\AblaExp}{
\begin{tabular}{llll}
    \toprule
    \textbf{} & \bf IS ↑         & \bf FID ↓         & \bf sFID ↓      \\
    \midrule
    Baseline  & 41.40           & 36.16             & 6.80             \\
    + $\Omega_{prot}$
              & 42.91\inc{1.51} & 35.01\decg{1.15}  & 6.72\decg{0.08}  \\
    + $\Omega_{dom}$  
              & 44.42\inc{3.02} & 33.94\decg{2.22}  & 6.53\decg{0.27}  \\
    \colorrow{}%
    + $\Omega_{prot}$ and $\Omega_{dom}$
              & 45.70\inc{4.30} & 32.75\decg{3.41}  & 6.56\decg{0.24}  \\
    \bottomrule
\end{tabular}
}

\newcommand{\ComponentExp}{
  \begin{tabular}{ccccc}
    \toprule
    \bf Param & \bf Value & \bf IS ↑ & \bf FID ↓ & \bf sFID ↓ \\
    \midrule
      & 10             & 45.01          & 33.14          & 6.63 \\
    \colorrow{}\cellcolor{white}
      & \textbf{12}    & \textbf{45.69} & \textbf{32.75} & \textbf{6.56} \\
      \multirow{-3}{*}{$\kappa$}
      & 16             & 44.94          & 33.21          & 6.63 \\
    \midrule
      & 0.02           & 44.19          & 34.07          & 6.68 \\
    \colorrow{}\cellcolor{white}
    \multirow{-2}{*}{$v_{\min}$}
      & \textbf{0.002} & \textbf{45.69} & \textbf{32.75} & \textbf{6.56} \\
    \bottomrule
  \end{tabular}
}

\newcommand{\REPA}{
\begin{tabular}{llllll}
    \toprule
    \textbf{} & \bf IS ↑ & \bf FID ↓ & \bf sFID ↓ & \bf Pre. ↑       & \bf Rec.  ↑  \\
    \midrule
    REPA~\citep{yu2024representation}  & 55.36 & 27.54 & 6.91 &0.56 &0.65 \\
    \colorrow{}%
    + Data Warmup~  & 58.08 \inc{2.72} & 25.84 \decg{1.70} & 6.89 \decg{0.02} &0.57 \inc{0.01}&	0.64\dec{0.01} \\
    \bottomrule
\end{tabular}
}

\newcommand{\imagenetSmall}{
\begin{tabular}{llll}
    \toprule
    \bf Model & \bf IS ↑ & \bf FID ↓  & \bf sFID ↓  \\
    \midrule
    SiT-S             & 28.02   & 98.31   & 221.72  \\
    \colorrow{} + Data Warmup & 29.09\inc{1.07} & 94.64\decg{3.67}  & 221.22\decg{0.5} \\
    SiT-B                & 43.05  & 79.38   & 220.69 \\
    \colorrow{} + Data Warmup & 28.75 & 100.97  & 225.38  \decg{4.69} \\
    SiT-L                & 52.41 & 76.36   & 226.27 \\
    \colorrow{} + Data Warmup & 21.30 & 127.51  & 241.50 \\
    \bottomrule
\end{tabular}
}

\newcommand{\modelScaling}{
\begin{tabular}{llll}
    \toprule
    \bf Model & \bf IS ↑ & \bf FID ↓ & \bf sFID ↓  \\
    \midrule
    SiT-S                  & 24.39 & 58.34 & 9.30  \\
    \colorrow{} + Warmup   & 25.55\inc{1.16} & 55.87\decg{2.47} & 9.16\decg{0.14}  \\
    \midrule
    SiT-B                  & 41.40 & 36.16 & 6.80  \\
    \colorrow{} + Warmup   & 45.70\inc{4.30} & 32.75\decg{3.41} & 6.56\decg{0.24} \\
    \midrule
    SiT-L                  & 71.48 & 18.79 & 5.14  \\
    \colorrow{} + Warmup   & 77.44\inc{5.96} & 17.17\decg{1.62} & 5.14\decg{0.00}  \\
    \midrule
    SiT-XL                 & 75.40 & 17.67 & 5.15  \\
    \colorrow{} + Warmup   & 80.07\inc{4.67} & 16.36\decg{1.31} & 5.13\decg{0.02}  \\
    \bottomrule
\end{tabular}
}

\newcommand{\datasetScaling}{
\begin{tabular}{lllll}
    \toprule
    \bf Dataset   & \bf Model      & \bf IS ↑ & \bf FID ↓ & \bf sFID ↓ \\
    \midrule
    & SiT-B/2        & 43.05           & 79.38            & 220.69           \\
    \colorrow{}\cellcolor{white}\multirow{-2}{*}{IN-100}
    & + Warmup   & 28.75\dec{14.30} & 100.97\incr{21.59} & 225.38\incr{4.69} \\
    \midrule
    & SiT-B/2        & 44.79           & 35.32            & 10.08            \\
    \colorrow{}\cellcolor{white}\multirow{-2}{*}{IN-500}
    & + Warmup   & 50.90\inc{6.11}  & 31.95\decg{3.37}  & 10.03\decg{0.05} \\
    \midrule
    & SiT-B/2        & 41.40           & 36.16            & 6.80             \\
    \colorrow{}\cellcolor{white}\multirow{-2}{*}{IN-1K}
    & + Warmup   & 45.70\inc{4.30}  & 32.75\decg{3.41}  & 6.56\decg{0.24}  \\
    \bottomrule
\end{tabular}
}
\begin{abstract}
A key inefficiency in diffusion training occurs when a randomly initialized network, lacking visual priors, encounters gradients from the full complexity spectrum—most of which it lacks the capacity to resolve. We propose Data Warmup, a curriculum strategy that schedules training images from simple to complex without modifying the model or loss. Each image is scored offline by a semantic-aware complexity metric combining foreground dominance (how much of the image salient objects occupy) and foreground typicality (how closely the salient content matches learned visual prototypes). A temperature-controlled sampler then prioritizes low-complexity images early and anneals toward uniform sampling. On ImageNet 256$\times$256 with SiT backbones (S/2 to XL/2), Data Warmup improves IS by up to 6.11 and FID by up to 3.41, reaching baseline quality tens of thousands of iterations earlier. Reversing the curriculum (exposing hard images first) degrades performance \emph{below} the uniform baseline, confirming that the simple-to-complex ordering itself drives the gains. The method combines with orthogonal accelerators such as REPA and requires only ${\sim}$10 minutes of one-time preprocessing with zero per-iteration overhead.
\end{abstract}   
\section{Introduction}
\label{sec:intro}

% --- P1: Concrete problem + hook ---
Training diffusion-based generative models~\citep{ho2020denoising, song2020score} is notoriously expensive: reaching high-fidelity synthesis~\citep{rombach2022high, saharia2022photorealistic} routinely costs hundreds of GPU-days. Much of this cost concentrates in the early optimization phase. A randomly initialized network, with no learned visual priors, must denoise images that range from a single centered object to cluttered multi-object scenes with complex backgrounds. Forcing such a model to reconstruct difficult images from the start is wasteful: the gradients are noisy and uninformative, contributing little to learning while consuming full compute.

% --- P2: Observation → Hypothesis → Challenge ---
This intuition suggests a remedy rooted in curriculum learning~\citep{bengio2009curriculum}: expose the model to simple images first, letting it build basic visual priors, then gradually introduce harder examples as its capacity grows. No drawing teacher starts with Picasso's Guernica. Yet applying this idea to diffusion training raises a concrete question: \emph{what makes an image ``simple'' for a diffusion model to learn?} Pixel-level statistics such as frequency content or compressibility are poor proxies; what matters is the semantic structure of the scene.

% --- P3: Our approach ---
We propose \textbf{Data Warmup}, a curriculum learning strategy that answers this question with a semantic-aware image complexity metric. The key design insight is that difficulty for a generative model correlates with two scene-level properties. First, \emph{foreground dominance}: an image whose frame is filled by a single salient object is easier to denoise than one where the same object appears small against a cluttered background. Second, \emph{foreground typicality}: a canonical, commonly seen view of an object is easier than an unusual angle or rare instance. We quantify both properties from pretrained DINO-v2 features in a single offline pass (Section~\ref{sec:image_complexity}) and combine them into a scalar score per image. A temperature-controlled sampler then uses these scores to bias early training toward low-complexity images, annealing toward uniform sampling over a warmup phase.

% --- P4: Preview of key findings ---
Empirically, the curriculum direction proves critical. On ImageNet-1K with SiT-B/2~\citep{ma2024sit}, Data Warmup substantially improves generation quality, while reversing the order (hard first) degrades both IS and FID below the uniform baseline. This rules out non-uniform sampling as the explanation: the simple-to-complex progression itself is the mechanism. Data Warmup further combines with the representation-alignment method REPA~\citep{yu2024representation} for additional gains, all for roughly ten minutes of one-time offline preprocessing.

% --- P5: Contributions ---
Our main contributions are summarized as follows:
\begin{itemize}
    \item We identify the mismatch between data complexity and model readiness as a concrete source of early-stage training inefficiency in diffusion models, and show that a simple-to-complex curriculum resolves it.
    \item We introduce a semantic-aware image complexity metric---combining foreground dominance and typicality---that requires only a single feature-extraction pass and drives a temperature-controlled sampling schedule.
    \item We demonstrate that Data Warmup improves IS by up to 6.11 and FID by up to 3.41 on ImageNet 256$\times$256 across SiT scales (S/2 to XL/2), complements existing accelerators like REPA, and reveals a sharp asymmetry: the same curriculum applied in reverse \emph{hurts} performance, establishing that ordering---not mere non-uniformity---is the mechanism.
\end{itemize}

\section{Related work}

Efforts to accelerate deep learning training generally follow three axes: structuring \emph{when} data is presented, selecting \emph{which data} to train on, and modifying \emph{how the model} processes that data.

\subsection{Curriculum learning}

Curriculum learning~\citep{bengio2009curriculum} formalizes the intuition that presenting training examples in a meaningful order---typically from easy to hard---can accelerate convergence and improve generalization. Early work demonstrated gains in language modeling and shape recognition by manually defining difficulty. Self-paced learning~\citep{kumar2010self} automates this by letting the model's own loss determine which samples are ``easy'', gradually raising a threshold to include harder ones. Subsequent extensions incorporate both self-paced and teacher-guided signals~\citep{jiang2015self}, or use reinforcement learning to select curricula~\citep{graves2017automated}.

Curriculum strategies have proven effective across domains: in machine translation~\citep{platanios2019competence}, where sentence length or rarity serves as a difficulty proxy; in reinforcement learning, where task complexity is staged~\citep{narvekar2020curriculum}; and in object detection, where training proceeds from clean to noisy annotations~\citep{wang2021survey}. However, most prior curricula rely on training-time signals (loss, gradient magnitude) that change every iteration, adding overhead and coupling the schedule to optimizer dynamics.

Data Warmup departs from this paradigm in two ways. First, difficulty is determined entirely \emph{offline} by a semantic complexity metric, decoupling the curriculum from training dynamics and introducing zero per-iteration cost. Second, we schedule via a temperature-controlled softmax rather than hard thresholding, ensuring a smooth transition that avoids abrupt distribution shifts.

\subsection{Data selection for efficient training}

A long line of work accelerates training by prioritizing informative samples~\citep{Wang_2026}. Classical importance-sampling methods~\citep{alain2015variance, katharopoulos2018not, loshchilov2015online, schaul2015prioritized} rank examples by training-time signals (gradient norms, loss values, or relevance to a validation set~\citep{mindermann2022prioritized}) and oversample the highest-scoring ones. Coreset and gradient-matching approaches~\citep{mirzasoleiman2020coresets, killamsetty2021grad} take a different approach, selecting minimal subsets that approximate the full gradient. These methods share a common limitation: they depend on signals that are themselves expensive to compute during training and whose theoretical guarantees break down in the non-convex regimes of modern deep networks, yielding diminishing returns at scale.

The closest precursor to our work is the prototype-based selection of \cite{lin2024prototypes}, which ranks images by their distance to cluster centroids and uses this ranking to accelerate Masked Autoencoder~\citep{he2021masked} pretraining. We borrow the idea of offline, feature-based scoring but depart in two key ways: (1)~we target generative diffusion training, whose learning dynamics differ fundamentally from masked reconstruction, and (2)~we introduce foreground dominance as an additional complexity axis, capturing the observation that images with prominent, centered objects are easier for a generative model to learn first.

\subsection{Accelerating diffusion model training}

Within generative modeling, most acceleration strategies are \emph{model-centric}: they modify architectures or training objectives. Architectural improvements include adapting transformer components for diffusion~\citep{wang2024fitv2} and redesigning timestep conditioning~\citep{yao2024fasterdit,ji2024advancing,ji2025translation}, while masked generative training reduces per-step computation by operating on partial inputs~\citep{zheng2023fast}. A second family leverages \emph{representation guidance}, injecting knowledge from pretrained encoders to bootstrap the denoising network. REPA~\citep{yu2024representation} aligns denoiser features with self-supervised vision encoders; SARA~\citep{chen2025sara} extends this to structural and distributional alignment; ERW~\citep{liu2025efficient} uses an external encoder to warm up internal representations.

All of these methods change the model or the loss; none change the order in which data is presented. Data Warmup is \emph{data-centric}: it leaves the model and objective untouched, modulating only the sampling distribution. Because it operates on a separate axis, it combines readily with model-centric accelerators; we show additive gains when stacking Data Warmup on top of REPA (Section~\ref{sec:main_exp}).

\section{Method}

\begin{figure*}[t]
    \centering
    \includegraphics[width = 0.83\textwidth]{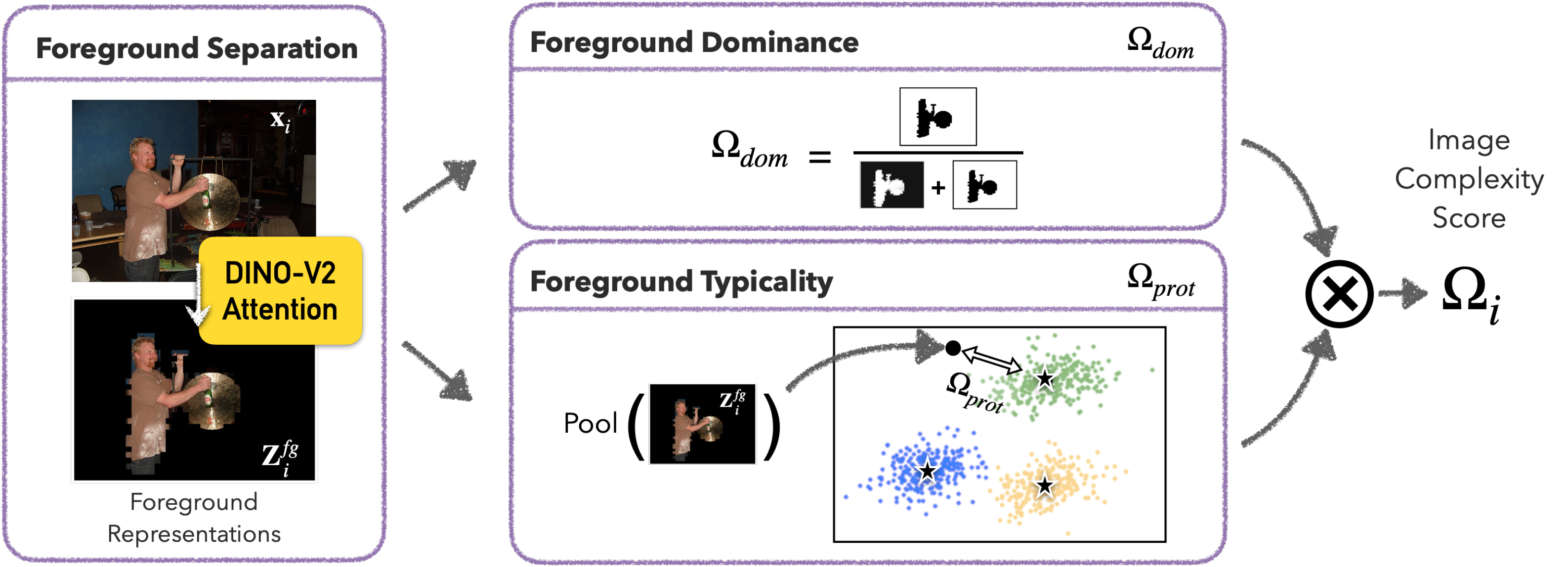}
    \caption{
        {\bf Image complexity.}
        Data warmup is based on image complexity. We quantify image complexity via two factors: foreground dominance and foreground typicality. Foreground separation leverages features from DINO-v2 to isolate salient regions. Foreground dominance ($\Omega_{dom}$) measures the prominence of foreground elements, while foreground typicality ($\Omega_{prot}$) assesses how representative a foreground is relative to learned prototypes. The overall complexity score ($\Omega_i$) is derived by combining these two factors.
    } %
    \label{fig:complexity}
\end{figure*}

\begin{figure*}[t]
    \centering
    \includegraphics[width = 0.85\textwidth]{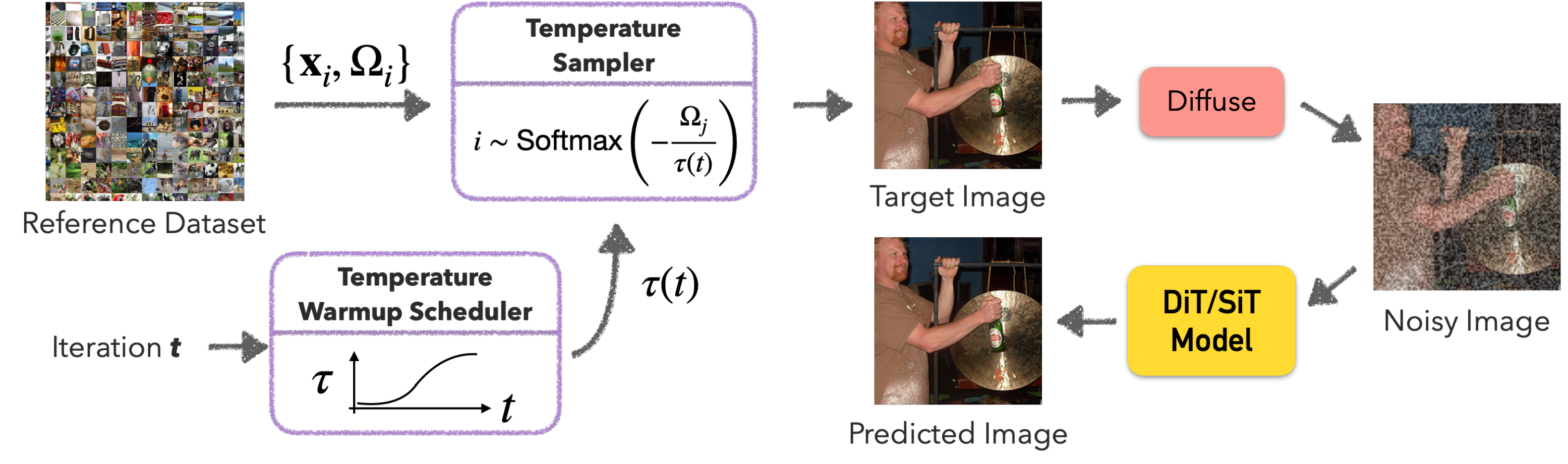}
    \caption{
        {\bf Data warmup.}
        During diffusion training, images are sampled according to their complexity scores through a temperature-based scheduler. Initially, simpler images (low complexity scores) are prioritized, with the sampling distribution gradually shifting to uniformly cover the full dataset complexity.
    } %
    \label{fig:warmup}
\end{figure*}

Data Warmup operates in two stages: an offline step that assigns each training image a scalar complexity score $\Omega$ (Section~\ref{sec:image_complexity}), and a temperature-controlled sampler that uses these scores to present images from simple to complex during training (Section~\ref{sec:dynamical_sampling}). Figures~\ref{fig:complexity} and~\ref{fig:warmup} illustrate them.

\subsection{Semantic-aware Image Complexity} %
\label{sec:image_complexity}

What makes an image easy or hard for a diffusion model to learn? Low-level statistics such as frequency content or compressibility are poor proxies: a cluttered photo and a textured close-up may have similar entropy yet pose very different challenges. We instead ground our complexity measure in the \emph{semantic structure} of the scene along two dimensions:
\begin{itemize}
    \item \emph{Foreground dominance} ($\Omega_{dom}$): how much of the image is occupied by salient objects. A centered golden retriever filling the frame is simpler than the same dog as a small figure in a busy park.
    \item \emph{Foreground typicality} ($\Omega_{prot}$): how representative the salient content is relative to visual prototypes. A canonical side-view photo of a dog is simpler than an unusual top-down angle of the same breed.
\end{itemize}
Our ablations (Section~\ref{tab:AblExp}) confirm that these two factors suffice; adding texture or compressibility features yields no further gain on curated datasets like ImageNet.

Both factors require identifying which parts of an image are foreground.

\subsubsection{Foreground-Background Separation} %
\label{subsec:fg_bg_sep}

We leverage DINO-v2~\citep{oquab2023dinov2}, whose spatial token representations correlate strongly with human visual saliency~\citep{yamamoto2024emergence}. For each image $i$, we extract the $L$ spatial token representations $\mathbf{z}_{i,1},\dots,\mathbf{z}_{i,L} \in \mathbb{R}^{d}$ and compute the first principal component $\mathbf{u}_1 \in \mathbb{R}^d$ across the dataset. Projecting each token onto $\mathbf{u}_1$ yields a per-location saliency score:
\begin{equation}
s_{i,j} = \mathbf{u}_1^\top \mathbf{z}_{i,j}, \quad j=1,\dots,L.
\end{equation}
Tokens scoring above a threshold $\theta$ are designated as foreground:
\begin{equation}
\mathbf{Z}_i^{fg} = \{ \mathbf{z}_{i,j} \mid s_{i,j} > \theta \}_{j=1}^L.
\label{eq:fg_detection}
\end{equation}
Following~\citep{oquab2023dinov2}, we set $\theta = 0.05$. The resulting set $\mathbf{Z}_i^{fg}$ serves as the basis for both complexity factors.

\subsubsection{Foreground Dominance $\Omega_\textit{dom}$}
\label{sec:fg_dominance}

The first factor measures how much of the image the foreground occupies. We define the background ratio $r_i^{bg} = (L - L_i^{fg}) / L$, where $L_i^{fg}=|\mathbf{Z}_i^{fg}|$.

\begin{figure}
    \centering
    % \vspace{-2em}
    \includegraphics[width=\linewidth]{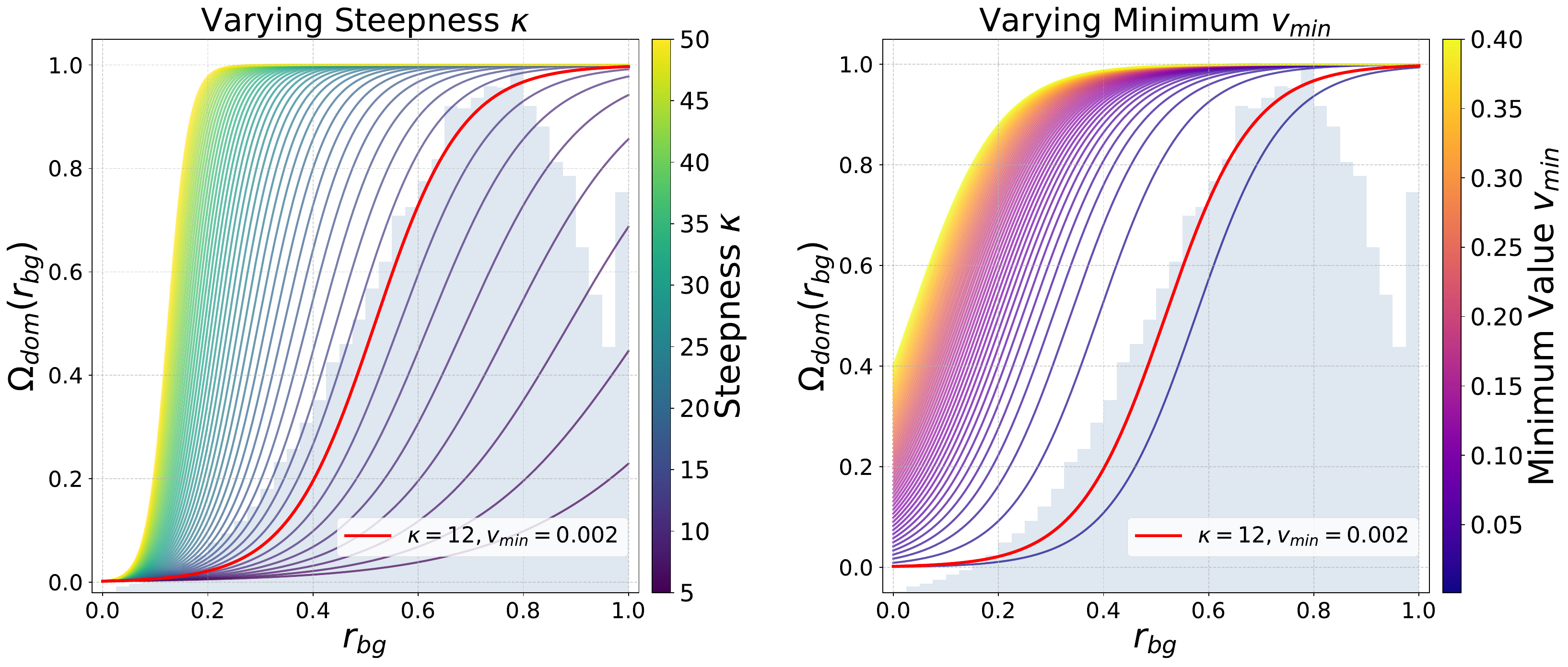} %
    \caption{Illustration of the complexity correction function $\Omega_{dom}(r_{bg}; \kappa, v_{\text{min}})$ (Eq.~\ref{eq:sigmoid_br}). \textbf{Left:} Varying steepness $\kappa$ with fixed $v_{\text{min}}=0.002$. Higher $\kappa$ leads to a sharper transition. \textbf{Right:} Varying minimum value $v_{\text{min}}$ with fixed steepness $\kappa=12.0$.}
    \label{fig:sigmoid_complexity}
\end{figure}

A linear mapping from $r_i^{bg}$ to complexity is a poor fit: the difference between 80\% and 60\% foreground coverage matters little for learning, whereas the drop from 40\% to 20\% signals a sharp increase in scene complexity. We capture this with a sigmoid correction:
\begin{equation}
\label{eq:sigmoid_br}
    \Omega_{dom}(\mathbf{z}_i) := \frac{1}{1 + e^{-(\kappa r_i^{bg} + \alpha(v_{\min}))}},
\end{equation}
$\text{where} \quad   \alpha(v_{\min}) = \ln\left(\frac{v_{\min}}{1-v_{\min}}\right),$
so that $\Omega_{dom} = v_{\min}$ when $r_i^{bg} = 0$. The steepness $\kappa$ controls how sharply complexity grows with background proportion, and $v_{\min}>0$ ensures that even fully foreground-dominated images retain a small non-zero score. We set $\kappa=12.0$ and $v_{\min}=0.002$ based on our hyperparameter study (Table~\ref{tab:hyperparam_ablation}). Figure~\ref{fig:sigmoid_complexity} visualizes the correction for varying $\kappa$ and $v_{\min}$.

\subsubsection{Foreground Typicality $\Omega_\textit{prot}$}
\label{sec:semantic_complexity}

Foreground size alone is insufficient: a large but unusual foreground (e.g., an extreme close-up of an insect wing) can still be hard to learn. Since prototypical images are generally easier to learn~\citep{lin2024prototypes}, the second factor captures how \emph{typical} the foreground content is relative to common visual patterns in the dataset.

We quantify typicality via clustering. For each image $i$, we average its foreground tokens into a single vector (using all tokens when no foreground is detected, a rare case on curated datasets like ImageNet):
\begin{equation}
    \label{eq:avg_foreground_rep}
    \bar{\mathbf{z}}_i^{fg} =
    \begin{cases} \frac{1}{L_i^{fg}} \sum_{\mathbf{z} \in \mathbf{Z}_i^{fg}} \mathbf{z} & \text{if } L_i^{fg} > 0 \\
    \frac{1}{L} \sum_{j=1}^L \mathbf{z}_{i,j} & \text{if } L_i^{fg} = 0. \end{cases}
\end{equation}
We apply $k$-means ($K{=}1000$ for ImageNet) to the full set $\{\bar{\mathbf{z}}_i^{fg}\}_{i=1}^N$, producing $K$ centroids $\{\boldsymbol{\mu}_k\}_{k=1}^K$ that serve as visual prototypes. Each image is assigned to its nearest centroid $k(i) = \arg\min_k \|\bar{\mathbf{z}}_i^{fg} - \boldsymbol{\mu}_k\|_2$, and its typicality score is the distance to that centroid:
\begin{equation}
    \label{eq:semantic_complexity_definition_fg}
    \Omega_{prot}(\mathbf{z}_i) := \| \bar{\mathbf{z}}_i^{fg} - \boldsymbol{\mu}_{k(i)} \|_2.
\end{equation}
Low values indicate prototypical foregrounds; high values indicate outliers.

\subsubsection{Overall Complexity Score}
\label{sec:overall_complexity}
\begin{figure}[t]
    \centering
    \includegraphics[width=0.95\linewidth]{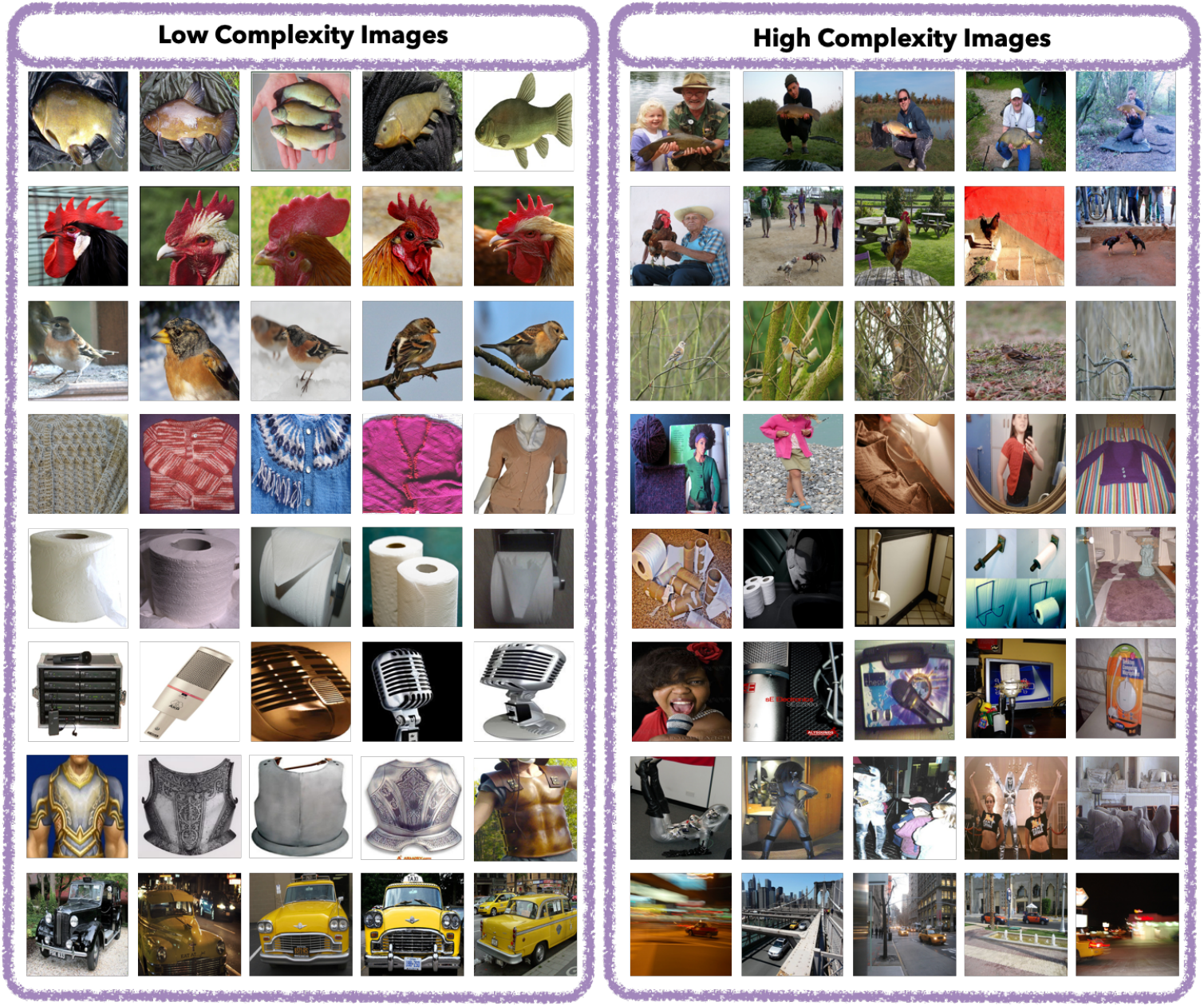}
    \caption{ImageNet samples organized into low and high complexity according to $\Omega$ scores. Each row displays images from the same category.}
    \label{fig:sample-complexity}
\end{figure}

An image should count as simple only when it is both foreground-dominated and prototypical, so we combine the two factors multiplicatively:
\begin{equation}
    \label{eq:overall_complexity}
    \Omega(\mathbf{z}_i) = \Omega_{dom}(\mathbf{z}_i) \times \Omega_{prot}(\mathbf{z}_i).
\end{equation}
A large but atypical foreground, or a prototypical object lost in clutter, each receive a high score. For brevity we write $\Omega_i = \Omega(\mathbf{z}_i)$ below. Figure~\ref{fig:sample-complexity} shows example images at both ends of the spectrum.

\subsection{Warmup Sampling Schedule}
\label{sec:dynamical_sampling}

We implement the curriculum as a temperature-controlled softmax over complexity scores, where the temperature rises over time.

\paragraph{Sampling probabilities.}
At training iteration $t$, image $i$ is sampled with probability
\begin{equation}
    P(i|t)
    = \frac{\exp\bigl(-\,\tilde \Omega_i / \tau(t)\bigr)}
       {\sum_{j=1}^N \exp\bigl(-\,\tilde \Omega_j / \tau(t)\bigr)},
    \label{eq:temp}
\end{equation}
where $N = |\mathcal{D}|$ is the number of training images. When $\tau$ is small, the distribution concentrates on low-$\Omega$ (simple) images; as $\tau \to \infty$, it flattens to uniform sampling.

However, the distribution of $\Omega_i$ varies substantially across prototype clusters, which would bias the curriculum toward visual concepts with lower raw scores\footnote{Similar biases were observed in \cite{lin2024prototypes}.}. We therefore normalize scores within each cluster before sampling:
\begin{equation}
    \tilde \Omega_i = \frac{\Omega_i - \Omega_{\min}^{k(i)}}{\Omega_{\max}^{k(i)} - \Omega_{\min}^{k(i)}},
\end{equation}
where $\Omega_{\min}^{k(i)}$ and $\Omega_{\max}^{k(i)}$ are the extremes within cluster $k(i)$. This ensures that at any temperature the sampler draws proportionally from all visual concepts, varying only the within-cluster difficulty.

\paragraph{Temperature annealing.}
Setting $\tau(t)$ directly is unintuitive. Instead, we parameterize the schedule through the \textbf{effective dataset size} $|{\cal D}_{\tau(t)}|$, defined as the expected number of unique images seen in one epoch under the current distribution~\citep{lin2024prototypes}:
\begin{equation}
    \label{eq:effective-size}
    |{\cal D}_{\tau(t)}| = \sum_{i=1}^{|{\cal D}|} \left[1 - \left(1 - P(i | t)\right)^{|{\cal D}|}\right].
\end{equation}

\begin{figure}
    \centering
    % \vspace{-1.1em}
    \includegraphics[width=0.8\linewidth]{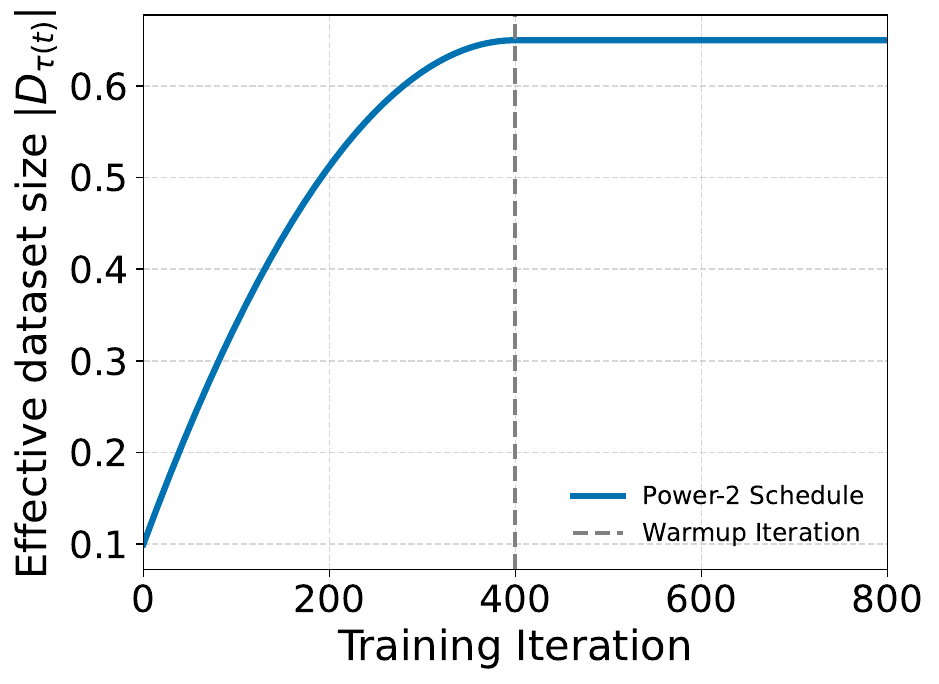} %
    \caption{
        {\bf Schedule.}
        Evolution of the effective dataset size $|{\cal D}_{\tau(t)}|$ (normalized by $|{\cal D}|$) using the power schedule. Warmup iterations set to $T_w=400$.
    }
    \label{fig:scheduler}
\end{figure}
Because $|{\cal D}_{\tau}|$ increases monotonically with $\tau$, the temperature corresponding to any target effective size can be recovered efficiently via binary search.

We schedule the effective dataset size to grow from an initial value $|{\cal D}_0|$ at $t=0$ to its theoretical maximum $|{\cal D}_{\text{max}}|=|{\cal D}| \times (1 - 1/e)$ at $t=T_w$, following a power-2 curve. The factor $1-1/e \approx 0.632$ is the expected fraction of distinct samples when drawing $|{\cal D}|$ times uniformly with replacement.
\begin{equation}
    |{\cal D}_{\tau(t)}| = |{\cal D}_0| + (|{\cal D}_{\text{max}}| - |{\cal D}_0|) \left(1 - \left[1 - \frac{t}{T_w}\right]_+^2\right),
\end{equation}
where $[\cdot]_+=\max(0, \cdot)$. This schedule (Figure~\ref{fig:scheduler}) accelerates rapidly through the simplest images and spends more iterations as the pool widens, since simple images need fewer exposures. For $t > T_w$, we switch to uniform sampling ($\tau \to \infty$).
Algorithm~\ref{alg:datawarmup} summarizes the full pipeline.

\begin{algorithm}[t]
\caption{Data Warmup}\label{alg:datawarmup}
\KwInput{Dataset $\mathcal{D}$, pretrained DINO-v2 encoder $\phi$, warmup iterations $T_w$, initial effective size $|{\cal D}_0|$, number of clusters $K$, sigmoid parameters $\kappa, v_{\min}$}
\BlankLine
\tcc{Stage 1: Offline complexity scoring}
\ForEach{image $i \in \mathcal{D}$}{
  Extract spatial tokens $\{\mathbf{z}_{i,j}\}_{j=1}^L \leftarrow \phi(i)$\;
  Compute saliency via PCA; partition tokens into $\mathbf{Z}_i^{fg}$ (Eq.~\ref{eq:fg_detection})\;
  $\Omega_{dom}(\mathbf{z}_i) \leftarrow \text{sigmoid}(\kappa, v_{\min}, r_i^{bg})$ \tcp*{Eq.~\ref{eq:sigmoid_br}}
  $\bar{\mathbf{z}}_i^{fg} \leftarrow \text{mean}(\mathbf{Z}_i^{fg})$ \tcp*{Eq.~\ref{eq:avg_foreground_rep}}
}
$\{\boldsymbol{\mu}_k\}_{k=1}^K \leftarrow k\text{-means}(\{\bar{\mathbf{z}}_i^{fg}\}_{i=1}^N)$\;
\ForEach{image $i \in \mathcal{D}$}{
  $\Omega_{prot}(\mathbf{z}_i) \leftarrow \|\bar{\mathbf{z}}_i^{fg} - \boldsymbol{\mu}_{k(i)}\|_2$ \tcp*{Eq.~\ref{eq:semantic_complexity_definition_fg}}
  $\Omega_i \leftarrow \Omega_{dom} \times \Omega_{prot}$; \quad normalize $\tilde\Omega_i$ within cluster $k(i)$\;
}
\BlankLine
\tcc{Stage 2: Curriculum training}
\For{$t = 1, 2, \dots, T_{\max}$}{
  \eIf{$t \leq T_w$}{
    Compute target $|{\cal D}_{\tau(t)}|$ via power-2 schedule\;
    Recover $\tau(t)$ by binary search\;
    Sample batch with $P(i|t) \propto \exp(-\tilde\Omega_i / \tau(t))$ \tcp*{Eq.~\ref{eq:temp}}
  }{
    Sample batch uniformly\;
  }
  Update model with standard diffusion loss\;
}
\end{algorithm}
\section{Experiments}

\paragraph{Setup.}
We train all models on ImageNet 256$\times$256~\citep{russakovsky2015imagenet} using the SiT framework~\citep{ma2024sit}, a generalized flow and diffusion-based architecture. Images undergo ADM preprocessing~\citep{dhariwal2021diffusion} and are encoded into latents $z \in \mathbb{R}^{32 \times 32 \times 4}$ via the Stable Diffusion VAE~\citep{rombach2022high}. We evaluate four backbone scales (SiT-S/2, B/2, L/2, XL/2) with 2$\times$2 non-overlapping patches and a batch size of 256. All warmup runs use a 200k-iteration curriculum phase followed by 200k iterations of uniform sampling; baselines train for 400k iterations with uniform sampling throughout. Training runs on A100 GPUs; preprocessing timing is reported on a single H100.

\paragraph{Complexity preprocessing.}
We extract DINO-v2 features in a single pass and run mini-batch $k$-means to obtain the $\Omega_{\text{prot}}$ clusters. For $\Omega_{\text{dom}}$, we threshold at 0.05 following~\citep{oquab2023dinov2}. The entire offline pipeline (feature extraction plus clustering) takes approximately ten minutes on one H100. At training time, only the softmax temperature in~\eqref{eq:temp} changes per iteration, adding negligible overhead.

\paragraph{Metrics.}
We report FID~\citep{heusel2017gans}, sFID~\citep{nash2021generating}, IS~\citep{salimans2016improved}, and precision/recall~\citep{kynkaanniemi2019improved}.

% ────────────────────────────────────────────────────────
\subsection{Does Direction Matter?}
\label{sec:main_exp}

We test whether the direction of the curriculum matters, or whether any non-uniform sampling would suffice. We compare three protocols on ImageNet-1K with SiT-B/2: (1)~uniform sampling (baseline), (2)~Data Warmup (simple$\to$complex), and (3)~\emph{inverse} warmup (complex$\to$simple). The inverse schedule uses exactly the same non-uniform sampling mechanism but reverses the direction, isolating ordering as the only variable.

Table~\ref{tab:MainExp} shows a clear asymmetry. Data Warmup improves IS by 4.30 and reduces FID by 3.41, while reversing the schedule \emph{actively degrades} performance (IS $-$4.80, FID $+$4.89), landing well below the uniform baseline. The gap between the two curricula ($\Delta$IS $\approx$ 9, $\Delta$FID $\approx$ 8) is far larger than either’s gap to the baseline, confirming that direction, not non-uniformity, is the key factor.

\paragraph{Compatibility with model-centric acceleration.}
Because Data Warmup operates solely on the data distribution, it should compose with methods that modify the model or loss. We verify this by stacking Data Warmup on top of REPA~\citep{yu2024representation}, a representation-alignment accelerator. As shown in Table~\ref{tab:RepaExp}, Data Warmup further improves REPA’s already strong results (IS $+$2.72, FID $-$1.70), suggesting that the two methods address different bottlenecks.

\begin{table}[t]
    \centering
    \small
    \setlength{\tabcolsep}{0.1pt}
    \caption{Performance of SiT-B/2 on ImageNet-1K (256$\times$256) with and without data warmup. The baseline employs uniform data sampling. ``Inverse Data Warmup’’ prioritizes high-complexity samples during early training phases. ↑/↓ denote that higher/lower is better. Superscripts denote performance gains/losses in green/red.}
    \MainExp
    % \vspace{4pt}
    \label{tab:MainExp}
\end{table}

\begin{table}[t]
    \centering
    \small
    \setlength{\tabcolsep}{1pt}
    \caption{Performance of REPA on ImageNet-1K (256×256) with and without data warmup using SiT-B/2 backbone. ↑/↓ denotes that higher/lower is better. Superscripts denote performance gains/losses in green/red.}
    \REPA
    % \vspace{4pt}

    \label{tab:RepaExp}
\end{table}

% ────────────────────────────────────────────────────────
\subsection{When Does Data Warmup Help---and When Does It Fail?}
\label{sec:scalability}

A curriculum that narrows the early training distribution necessarily trades diversity for focus. When is this trade-off beneficial, and when does it backfire? We probe two axes, dataset size and model capacity, to identify when it helps.

\begin{table}[t!]
\small
  \centering
    \caption{SiT-B/2 evaluated on datasets of increasing size. ↑/↓ denotes that higher/lower is better. Superscripts denote performance gains/losses in green/red.}
    \datasetScaling
    % \vspace{4pt}

    \label{tab:in100}
\end{table}

\paragraph{Dataset size: a diversity threshold.}
We train SiT-B/2 on three ImageNet subsets of increasing size: IN-100 (${\sim}$100 images/class), IN-500 (${\sim}$500), and IN-1K (${\sim}$1{,}000). The results (Table~\ref{tab:in100}) reveal a clear threshold effect. On IN-100, Data Warmup \emph{hurts}: IS drops by 14.30 and FID worsens by 21.59. The dataset is simply too small; concentrating early sampling on ``simple’’ images starves the model of the diversity it needs to learn the full distribution, causing it to overfit to a narrow manifold of canonical examples.

The picture reverses sharply at IN-500 and above. On IN-500, warmup improves IS by 6.11 and FID by 3.37, the largest gains in our experiments. On IN-1K, the improvements remain strong (IS $+$4.30, FID $-$3.41). The takeaway: Data Warmup requires enough diversity that focusing on simple samples early does not starve the model, a condition easily met by any practical large-scale dataset.

\begin{figure}[t]
    \centering
    % \vspace{-1em}
    \includegraphics[width=0.7\linewidth]{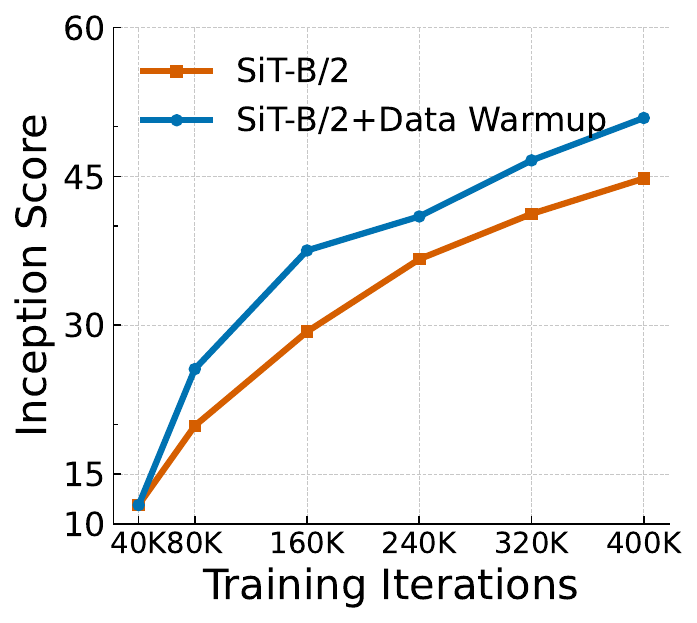} %
    \caption{{\bf Training curve.} Inception scores of an SiT model trained on ImageNet-500 with and without data warmup. $T_w=200k$}
    \label{fig:train-curve}
\end{figure}

Figure~\ref{fig:train-curve} visualizes the dynamics on IN-500. The warmed-up model pulls ahead from the earliest iterations, with the gap widening through the curriculum phase (40k--160k iterations) and persisting at 400k iterations with a lead of approximately 6 IS points. Data Warmup thus accelerates convergence \emph{and} improves final quality; it is not merely ``front-loading’’ gains that the baseline eventually recovers.

\paragraph{Model capacity: consistent gains across scales.}
Table~\ref{tab:in1k} evaluates four SiT backbones on ImageNet-1K. Data Warmup improves every model, with gains that grow with capacity: IS improvements range from $+$1.16 (SiT-S/2) to $+$5.96 (SiT-L/2). Larger models show bigger gains, possibly because they have more parameters to benefit from structured early gradients; smaller models converge faster and spend less time in the regime where the curriculum matters most.

% ────────────────────────────────────────────────────────
\subsection{Are Both Complexity Factors Necessary?}
\label{sec:ablation}

We ablate the two components of the complexity score to understand their individual and joint contributions, then verify sensitivity to the sigmoid hyperparameters in $\Omega_{dom}$.

\paragraph{Individual vs.\ combined factors.}
Table~\ref{tab:AblExp} compares curricula driven by $\Omega_{dom}$, $\Omega_{prot}$, and their product. Each factor independently improves over uniform sampling, with foreground dominance contributing the larger share (IS $+$3.02 vs.\ $+$1.51). Combining them yields gains (IS $+$4.30, FID $-$3.41) that exceed the sum of the individual improvements, indicating a synergy: an image can have a dominant foreground yet be atypical, or be prototypical yet cluttered. The multiplicative score penalizes complexity along either axis.

\begin{table}[t]
    \centering
    \caption{SiT backbones of varying size evaluated on ImageNet 1000 (256×256). ↑/↓ denotes that higher/lower is better. Superscripts denote performance gains/losses in green/red.}
    \scalebox{0.8}{\modelScaling}
    % \vspace{4pt}

    \label{tab:in1k}
\end{table}

\begin{table}[h]
        \centering
        \caption{Performance comparison of SiT training strategies using the SiT-B/2 model on ImageNet-1K (256$\times$256). ↑/↓ denotes that higher/lower is better. Superscripts denote performance gains/losses in green/red.}%
        \AblaExp
        % \vspace{4pt}

        \label{tab:AblExp}
\end{table}
\begin{table}[h]
        \centering
        \caption{Study of $\Omega_{dom}$ hyper-parameters $\kappa$ and $v_{min}$, evaluated on SiT-B/2 (256×256). The impact of $\kappa$ and $v_{min}$ is evaluated around the optimal setting $\kappa=12, v_{min}=0.002$.}
        \ComponentExp
        % \vspace{4pt}

        \label{tab:hyperparam_ablation}
\end{table}

\begin{figure*}[t]
    \centering
    \includegraphics[width=1\textwidth]{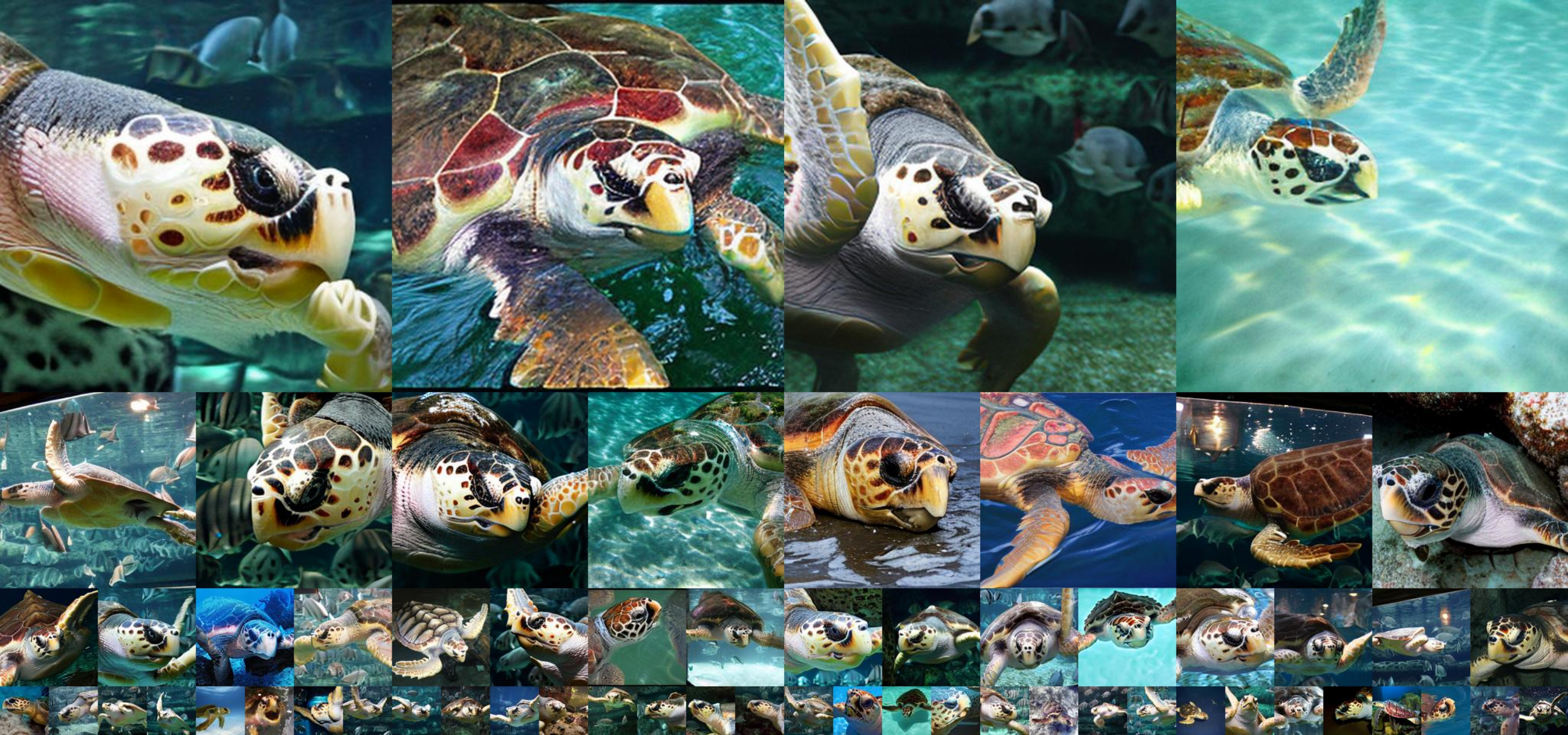}
    \caption{Generated samples for the class ``Loggerhead Sea Turtle’’ from a SiT-XL/2 model trained with REPA and data warmup for 2M iterations (classifier-free guidance $w=4.0$).}
    \label{fig:qualitative_turtles}
\end{figure*}

\paragraph{Sigmoid hyperparameters.}
Table~\ref{tab:hyperparam_ablation} varies the steepness $\kappa$ and floor $v_{\min}$ of the sigmoid correction in $\Omega_{dom}$. Performance is robust within a reasonable range: $\kappa \in [10,16]$ and $v_{\min} \in [0.002, 0.02]$ all outperform the baseline, with $\kappa=12$, $v_{\min}=0.002$ achieving the best overall results.

% ────────────────────────────────────────────────────────
\subsection{Qualitative Examples}

To assess generation quality beyond aggregate metrics, we train a SiT-XL/2 model with REPA and Data Warmup for 2M iterations (classifier-free guidance $w=4.0$). Figure~\ref{fig:qualitative_turtles} shows samples for the class ``Loggerhead Sea Turtle.’’ The model reproduces fine-grained details (shell scute patterns, skin textures, light caustics underwater) across diverse poses and environments, consistent with the hypothesis that easy-first curricula help establish structural priors early.

\section{Discussion}

\textbf{Why does direction matter so strongly?}
The most striking result in our experiments is not that Data Warmup helps, but that \emph{reversing} it actively hurts---landing well below even the uniform-sampling baseline (Section~\ref{sec:main_exp}). We hypothesize that simple, foreground-dominated images provide a low-entropy gradient signal that guides the randomly initialized network toward a structured region of parameter space early on. Once these foundational priors are established, the model can meaningfully learn from harder, more ambiguous scenes. The inverse schedule does the opposite: it floods the blank-slate model with high-complexity images whose gradients conflict and average out, pushing the model into a flatter, less informative loss region from which recovery is slow. This interpretation aligns with recent analyses of loss landscape geometry in diffusion models~\citep{yao2024fasterdit} and suggests that the early training phase has an outsized, potentially irreversible influence on the trajectory of optimization.

\noindent\textbf{The focus--diversity trade-off.}
The IN-100 failure (Section~\ref{sec:scalability}) reveals a design principle that extends beyond our specific method: \emph{any curriculum that narrows the training distribution must be backed by sufficient data diversity to avoid mode collapse}. On IN-100 (${\sim}$130k images), concentrating early sampling on simple examples leaves too few unique images per epoch, causing the model to overfit to a narrow manifold. On IN-500 and above, the dataset is rich enough that even a focused curriculum still exposes the model to substantial within-class variation. This threshold effect suggests that Data Warmup is naturally suited to the large-scale regimes where training cost is highest and acceleration is most valuable.

\noindent\textbf{Limitations and future directions.}
Our complexity metric is computed offline from a frozen DINO-v2 backbone and remains static throughout training. An adaptive, loss-aware variant that re-scores images as the model improves could yield tighter curricula, especially in later training phases where the current schedule defaults to uniform sampling. More broadly, we have only evaluated Data Warmup on class-conditional image generation. Extending the framework to text-conditioned models---where \emph{prompt complexity} introduces an entirely new curriculum axis (``a dog'' vs.\ ``a golden retriever playing fetch in a sunlit park with children'')---is a natural and promising direction. Finally, while our two-factor complexity metric suffices for curated datasets like ImageNet, richer metrics may be needed for uncurated, web-scale data where texture, occlusion, and label noise introduce additional difficulty dimensions.

\section{Conclusion}

This study identifies that training diffusion models uniformly over the full data distribution from the start is wasteful: the randomly initialized network cannot learn from complex scenes it does not yet understand. Data Warmup resolves this mismatch with a simple curriculum---score each image by foreground dominance and typicality, then anneal a temperature-controlled sampler from easy to hard. On ImageNet-256 with SiT backbones (S/2 to XL/2), this improves IS by up to 6.11 and FID by up to 3.41, reaching baseline quality tens of thousands of iterations earlier. The method stacks with REPA for further gains (IS $+$2.72, FID $-$1.70) and costs only ${\sim}$10 minutes of offline preprocessing with zero per-iteration overhead. Perhaps most importantly, reversing the curriculum \emph{hurts}---establishing that what matters is not non-uniform sampling per se, but the simple-to-complex ordering that lets a model build structure before confronting complexity.

{
    \small
    \bibliographystyle{ieeenat_fullname.bst}
    \bibliography{main.bib}
}


\end{document}